\documentclass[letterpaper, 10 pt, conference]{ieeeconf}  

\IEEEoverridecommandlockouts                              

\usepackage{graphics} 
\usepackage{epsfig} 
\usepackage{amsmath} 
\usepackage{amsfonts}
\usepackage{xcolor}
\usepackage{subfigure}
\usepackage{booktabs}
\usepackage{multirow}

\title{\LARGE \bf
Mode-GS: Monocular Depth Guided Anchored 3D Gaussian Splatting for Robust Ground-View Scene Rendering 
}

\author{Yonghan Lee$^{1,2}$, Jaehoon Choi$^{1}$, Dongki Jung$^{1}$, Jaeseong Yun$^{2}$, \\ Soohyun Ryu$^{2}$,  Dinesh Manocha$^{1}$,  and Suyong Yeon$^{2}$ 
\thanks{$^{1}$University of Maryland,
        8125 Paint Branch Dr, College Park, MD 20742, USA.}
\thanks{$^{2}$NAVER LABS, 95 Jeongjail-ro, Bundang-gu, Seongnam-si, Gyeonggi-do, South Korea.}%
}

\begin{document}

\maketitle
\thispagestyle{empty}
\pagestyle{empty}

\begin{abstract}

We present a novel-view rendering algorithm, Mode-GS, for ground-robot trajectory datasets. Our approach is based on using anchored Gaussian splats, which are designed to overcome the limitations of 
existing 3D Gaussian splatting algorithms.  Prior neural rendering methods suffer from severe splat drift due to scene complexity and insufficient multi-view observation, and can fail to fix splats on the true geometry in ground-robot datasets. 
Our method integrates pixel-aligned anchors from monocular depths and generates Gaussian splats around these anchors using residual-form Gaussian decoders. To address the inherent scale ambiguity of monocular depth, we parameterize anchors with per-view depth-scales and employ scale-consistent depth loss for online scale calibration. Our method results in improved rendering performance, based on PSNR, SSIM, and LPIPS metrics, in ground scenes with free trajectory patterns, and achieves state-of-the-art rendering performance on the R$^{3}$LIVE odometry dataset and the Tanks and Temples dataset. 


\end{abstract}

\section{Introduction}




The development of navigation and perception algorithms for autonomous robots typically requires extensive and costly field experiments for training and validation.
In this context, neural rendering offers a practical solution, as it can significantly reduce the time and effort based on data simulation and augmentation~\cite{nerf-nav, splat-nav, maxey2023uav}. Specifically, neural rendering can be used to generate novel view images by learning a neural scene representation from a set of input training images and corresponding poses~\cite{mildenhall2021nerf, kerbl20233d}. 

Current neural rendering research is based on two popular methods: \textit{implicit} Neural Radiance Fields (NeRF) \cite{mildenhall2021nerf} and \textit{explicit} 3D Gaussian Splatting (3DGS) \cite{kerbl20233d, zwicker2002ewa}. NeRF gained popularity because of its high-fidelity and continuous scene representations due to the implicit nature of radiance fields. However, it is hard to scale to large scenes because of the limited representational capacity of its coordinate-based Multi-Layer Perceptron (MLP), which cannot efficiently handle the cubic growth in scene complexity. On the other hand, 3DGS provides a feasible alternative for scene-scale rendering of ground-view robot datasets by explicitly representing only the non-empty parts of the scene, and using more interpretable Gaussian splats as scene primitives. 
\begin{figure}[t]
    \centering
    \includegraphics[width=\linewidth]{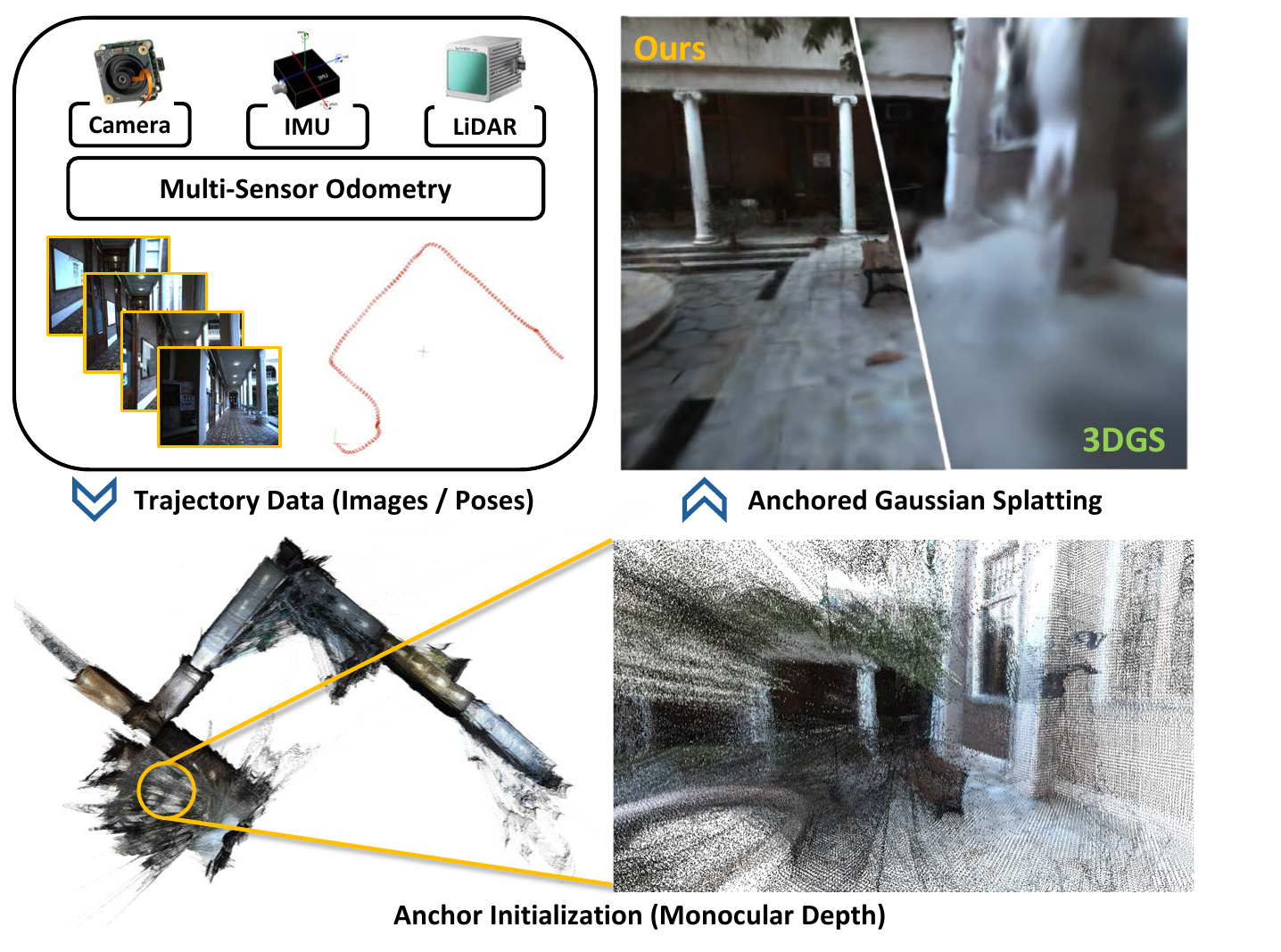}
    \caption{Our Mode-GS integrates \textit{monocular depth estimation} with \textit{anchored Gaussian splatting}, uses a scale-consistent depth calibration technique and residual-based Gaussian decoders. By incorporating dense pixel-aligned anchor points from monocular depth, anchored splatting improves robustness in scenarios without dense multi-view images and mitigates the impact of inaccurate poses in complex ground-view scenes. Our method can be developed using multi-sensor odometry poses in a point-cloud-free setting. Overall, it offers a practical and robust rendering pipeline for ground-view robotic datasets, as shown in Section V.}
    \label{fig:intro_figure}
\end{figure}
 It turns out that complex ground-view robot datasets with free trajectories~\cite{wang2023f2nerf} present two fundamental challenges with respect 3DGS algorithms: (1) the scarcity of multiple-view information; (2) the difficulty of obtaining pixel-level accurate trajectory poses \cite{Colmap-free3DGS,liu2024gsloc,zhao2024badgaussians}.

First of all, 3DGS requires dense point clouds for splat initialization and informative multi-view photometric gradients for Adaptive Density Control (ADC)~\cite{kerbl20233d} to expand splats into unoccupied areas, which significantly deteriorates its performance on ground-robot datasets. Since training a neural rendering algorithm heavily relies on dense multi-view observations due to its inherent high dimensionality, previous neural rendering approaches have largely focused on datasets with structured viewing patterns---such as aerial (top-down view)~\cite{Turki2022meganerf,xiangli2022bungeenerf},object-centric (inward view)~\cite{barron2022mipnerf360,Knapitsch2017tnt}, and street (forward motion) datasets~\cite{Liao2022Kitti360}. However, there is relatively less work on ground-robot datasets with free trajectories~\cite{lin2022r3live}.

The second challenge in ground-view rendering is the sensitivity of 3DGS to pixel-level pose accuracy of training images, which is crucial due to its reliance on pixel-based photometric loss. Acquiring pixel-accurate poses in ground-view datasets is difficult. Structure from Motion (SfM)~\cite{COLMAP} or vision-based SLAM methods~\cite{ORBSLAM3_TRO} often struggle to consistently estimate poses in ground-robot datasets without fragmenting or diverging, when images lack salient features or textures to track. While multi-sensor SLAM and odometry methods~\cite{legoloam2018shan, qin2017vins, lin2022r3live} are more reliable for trajectory pose estimation, they frequently fail to achieve pixel-level accuracy due to heterogeneous sensor configurations and sensor fusion. Due to sensitivity of 3DGS on pose accuracy, the SLAM algorithms that directly integrate 3D Gaussian splats~\cite{MonoGS, keetha2024splatam, yan2023gsslam} as scene representation often fail to reliably estimate complete trajectories.


\begin{figure}[t]
    \centering
    \includegraphics[width=1.0\linewidth]{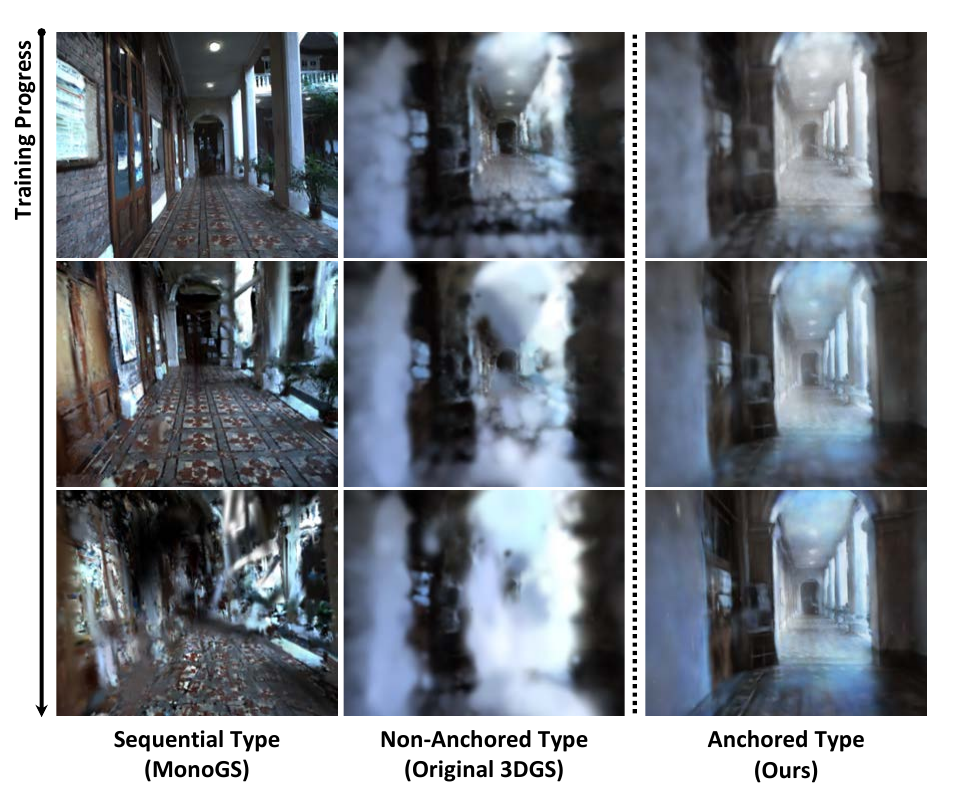}
    \vspace*{-5mm}
    \caption{ 
    We compare the degenerate training patterns of 3DGS in scenarios without dense multi-view information. The patterns are categorized according to their type:
    (a) \textbf{Sequential Type}: SLAM-based Gaussian splatting utilizes sequential information by processing consecutive images with pose refinement, initially generating sharper images. However, their pose tends to drift and eventually diverges;
    (b) \textbf{Non-Anchored Type}: In the original 3DGS and their variants with ADC, the splats tend to \textbf{drift} from the true geometry without dense multi-view photometric information;
    (c) \textbf{Anchored Type}: Anchoring effectively prevents splats from becoming detached from the actual geometry.
    } 
    \label{fig:degen_comparison}
\end{figure}

\noindent {\bf Main Results:} We present a novel rendering approach, Mode-GS, to address these challenges in ground-robot datasets. Our method integrates monocular depth networks with an anchored Gaussian splat generation, incorporating a scale-consistent depth calibration mechanism. By utilizing monocular depths, we initialize pixel-aligned anchors that fully cover the frustums across all input images, effectively preventing drift of splats caused by degenerate densification in the absence of sufficient multi-view photometric cues (Fig. \ref{fig:degen_comparison}). Being inspired by \cite{lu2024scaffoldgs}, we design \textit{Residual-form Gaussian Decoder} to robustly generate Gaussian splats around these anchors. Unlike the decoder structure in~\cite{lu2024scaffoldgs}, our novel decoder enables direct initialization of splat attributes (e.g. color) and greatly improves the training efficiency  thanks to the efficient residual structure. Lastly, the inherent scale ambiguity of monocular depth is mitigated during the training by our \textit{Anchor Depth-Scale Parameterization} and \textit{Scale-Consistent Depth Loss}, leading to the consistent depth calibration. Our novel contributions include:


\begin{itemize} 
\item \textbf{Scale-consistent Integration of Monocular Depths}: We introduce a novel approach that integrates scale-ambiguous monocular depth networks to initialize pixel-aligned splats. Through our {anchor depth-scale parameterization} and {scale-consistent depth loss}, we achieve consistent depth calibration during training, eliminating the need for initial SfM or LiDAR point clouds.

\item \textbf{Anchor-Decoder Structure with Residual-Form MLP Decoder}: We present an advanced anchor-decoder structure for 3DGS, featuring our proposed residual-form Gaussian decoder. This allows for the direct initialization of anchored Gaussian splat attributes, improving both the efficiency and accuracy of the scene training process.  

\item \textbf{Novel-view-synthesis from Ground-Robot Dataset}: Our method shows robust rendering performance on ground-robot datasets with free trajectory patterns, achieving state-of-the-art rendering performance on the R$^{3}$LIVE odometry dataset \cite{lin2022r3live} even without LiDAR point clouds, while maintaining comparable performance on the Tanks and Temples dataset. Note that our algorithm can be built on easily obtainable odometry poses in a point-cloud free setting, providing practical and robust rendering pipeline for ground-view robot datasets. We highlight the improvements over prior methods in terms of rendering metrics (PSNR, SSIM, and LPIPS)  on these datasets in Section V.


\end{itemize}

\section{Related Works}

\noindent\textbf{Neural Rendering for Robot Navigation} 
Neural rendering \cite{mildenhall2021nerf,kerbl20233d} can achieve remarkable photo-realistic rendering quality. This photorealistic rendering can be applied to various robotics applications such as navigation \cite{nerf-nav,splat-nav,liu2024integrating}, robot data simulation \cite{maxey2023uav,meyer2024pegasus}, and robotic teleoperation \cite{patil2024radiance}. Prior works present the robot navigation methods designed for 3D environment represented as NeRF \cite{nerf-nav,liu2024integrating} or 3D Gaussian Splats \cite{splat-nav}. UAV-Sim \cite{maxey2023uav} and PEGASUS \cite{meyer2024pegasus} utilize neural rendering to synthesize data for aerial perception or robotic manipulation.       

\vspace{0.5mm}
\noindent\textbf{3D Gaussian Splatting with Geometric Prior}
3D Gaussian Splatting (3DGS)~\cite{kerbl20233d} has received considerable attention. 3DGS utilizes 3D Gaussian splats \cite{zwicker2002ewa} as 3D primitives which can be rendered thorough sorting and rasterization. Scaffold-GS~\cite{lu2024scaffoldgs} introduces anchors to cluster adjacent 3D Gaussians and uses an Multi-Layer Perceptron (MLP) to predict their attributes. However, since 3D Gaussian splats rely solely on photometric constraints, they often violate geometric coherence. Several studies~\cite{sugar,chen2023neusg} focusing on surface extraction from Gaussian splats have addressed this issue by aligning the flat 3D Gaussian splats with the geometry. 2DGS ~\cite{2dgs} directly projects 3D Gaussian onto flat 2D Gaussian for effective mesh extraction. Gaussian Opacity Fields ~\cite{gof} extract surfaces using ray-tracing based volume rendering. To consider geometry regularization, prior methods either used monocular depth estimation \cite{radegs,dnsplatter} or multi-view stereo \cite{cheng2024gaussianpro,mvgsplatting} for designing training strategies. 

\vspace{0.5mm}
\noindent\textbf{3D Gaussian Splatting with SLAM}  
All of previous methods depend on precise poses obtained from a Structure from Motion system~\cite{COLMAP}, which require significant computational resources.
Recent research has expanded the 3DGS by incorporating various SLAM systems, including those using monocular camera \cite{MonoGS} or RGB-D sensors \cite{keetha2024splatam}.The aforementioned methods are limited to small-scale scenes that allow for the collection of dense viewpoints and cover large areas of the environment. Additionally, CF-3DGS \cite{Colmap-free3DGS} and InstantSplat \cite{fan2024instantsplat} present end-to-end frameworks for joint novel view synthesis and camera pose estimation from sequential images. However, none of these methods effectively estimate a reliable camera trajectory in ground-view robot datasets for training the Gaussian Splatting technique.



\begin{figure*}[t]
    \centering
    \includegraphics[width=1.0\linewidth]{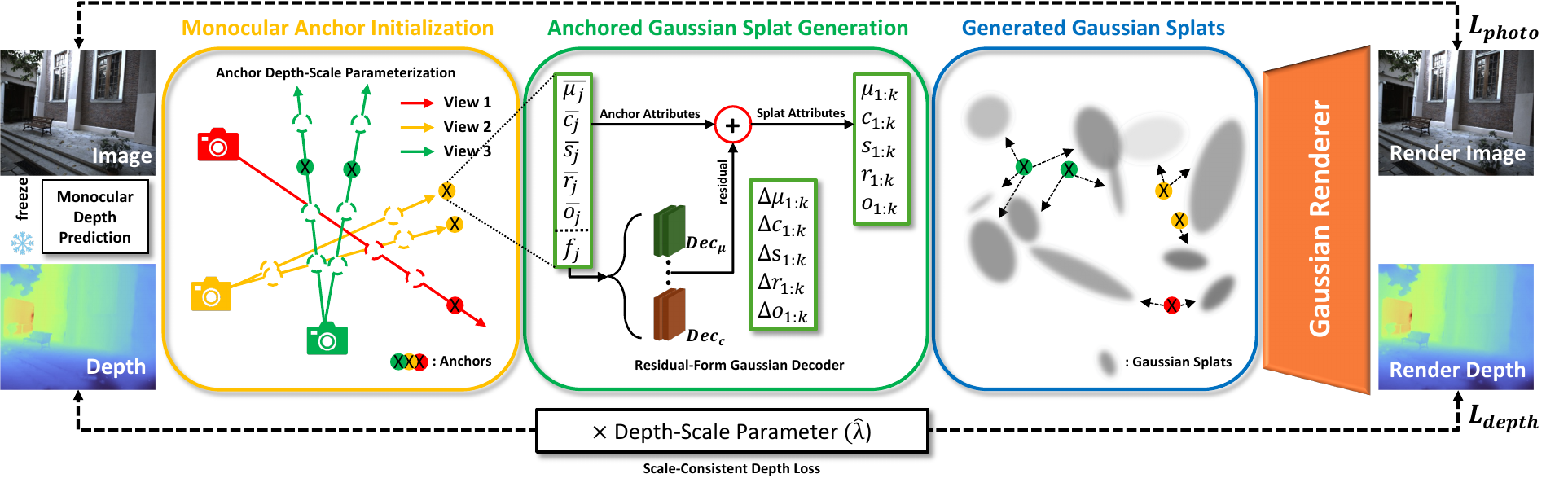}
    \vspace*{-5mm}
    \caption{ 
    Our methods consists of three main steps:
    (a) \textbf{Per-View Anchor Initialization}: Given monocular depth images, depth-scale adjustable anchors are initialized from each view. Each anchor is fixed in the 3D scene except the depth-scale toward the corresponding view.
    (b) \textbf{Anchor Decoding with Residual-Form Gaussian Decoder}: Each anchor is decoded into $k$ Gaussian splats by our residual-form Gaussian decoders. When initialized, each anchor contains nominal Gaussian splat attributes $(\bar{\mu}_j, \bar{r}_j, \bar{c}_j, \bar{o}_j, \bar{s}_j)$ and an embedded feature $f_j$. The residual decoders generate $k$ sets of residual attributes for child splats, which are combined with nominal anchor attributes to generate child Gaussian splats.
    (c) \textbf{Training with Scale-Consistent Depth Loss} Online Depth-Scale Calibration: We use scale-consistent depth loss $\mathcal{L}_\text{depth}$
that incorporates scales for each monocular depth supervision.}
    \label{fig:pipeline}
\end{figure*}

\section{Preliminary}
3DGS \cite{kerbl20233d} is a differentiable rendering method, which can be trained from $n$ images $\{I_1, I_2, \cdots, I_n\}$ to learn volumetric scene representation $\mathcal{G}$ as a mixture of anisotropic Gaussian primitives $\{G_1, G_2, \cdots, G_m\}$. From the trained scene representation $\mathcal{G}$, tile-based differentiable rasterizer $\mathcal{R}$ can render image $I$ for novel view $T \in \text{SE(3)}$ as $I = \mathcal{R} (\mathcal{G} \,|  \,T)$.

Gaussian primitives ${G_i}$ are initialized from input point cloud with means $\mu_i$ as the corresponding point positions. As such, each Gaussian $G(x)$ is represented as,

\begin{equation}
G(x) = \exp\left({-\frac{1}{2}(x - \mu)^T \Sigma ^{-1} (x - \mu)}\right)
\end{equation}

where x is arbitrary position in 3D space and $\Sigma$ denotes covariance of Gaussian kernel. Here, $\Sigma$ is decomposed into scaling matrix $S$ and rotation matrix $R$ to preserve semi positive-definite form, $\Sigma = RSS^TR^T$
Additionally, each Gaussian stores an opacity value $o_i$, which is multiplied with $G(x)$ as $\alpha_i(x) = G_i(x) o_i $ for alpha-blending weight $\alpha_i$, and the view-dependent color $c_{i}$ represented by spherical harmonics (SH). 3D Gaussians are projected into 2D space \cite{zwicker2002ewa}, and the color $C$ is computed through volumetric rendering, following a front-to-back depth order:
\begin{equation}
C = \sum_{i \in N} c_i \alpha_i \prod ^{i-1}_{j=1} (1 - \alpha_i)
\end{equation}
where $N$ is the set of sorted Gaussians overlapping with the given pixel.

\section{Methods}

Our method aims to enabling the rendering of novel view images from ground-view robot trajectory datasets, which is challenging for the original 3DGS algorithms due to complexity of ground-view scenes and lack of multi-view observations with free trajectory pattern (Fig. \ref{fig:degen_comparison}). Our ground-view rendering pipeline integrates monocular depth networks \cite{yin2023metric} into anchored Gaussian splatting \cite{lu2024scaffoldgs} scheme with the proposed scale-consistent depth calibration framework. This approach leverages pixel-aligned splat initialization from monocular depth networks, thereby further improving the robustness of the anchored Gaussian structure \cite{lu2024scaffoldgs}.

As can be seen in Fig. \ref{fig:pipeline}, our pipeline consists of three main steps: 1) \textit{Per-View Anchor Initialization}, 2) \textit{Anchored Gaussian Splat Generation}, and 3) \textit{Training from Rendering Losses}. Instead of directly initializing Gaussian splats, we initialize anchor points with embedded features and nominal Gaussian attributes, and generate child Gaussian splats by combining the nominal attributes with the residuals decoded from the Gaussian decoders. Note that our pipeline does not require initial SfM or LiDAR point clouds, but built on easy-to-obtain odometry poses and corresponding images, which increases the practicality of our approach for real-world datasets.

\subsection{Per-View Anchor Initialization}

\noindent\textbf{Monocular Anchor Initialization}
Instead of utilizing input SfM or LiDAR point clouds, we utilize monocular depth networks to initialize anchor points $\mathcal{P} $ fused from each training view $i$. Firstly, monocular depth images \cite{yin2023metric} $\{D_i\}_{1:n}$ are generated from training images $\{I_i\}_{1:n}$. With given odometry poses $\{(R_i, t_i)\}_{1:n}$ with $R_i \in \text{SO(3)}$, $t_i \in \mathbb{R}^3$. depth images $D_{1:n}$ are unprojected to the 3D space. The 3D points are then voxelized with small resolution $\epsilon$ to remove redundant anchor points, generating a cluster of per-view anchor points $\mathcal{P} = \{\mathcal{P}_i\}_{i \in 1:n}$.

\vspace{0.5mm}
\noindent\textbf{Anchor Depth-Scale Parameterization} 
Each set of per-view anchor points $\mathcal{P}_i$ has inherent scale-ambiguity which needs to be calibrated for multi-view consistency. To allow depth-scale adjustment of each per-view anchor points $\mathcal{P}_i$, we introduce scale parameter $\hat{s}_i$ for each view $i$. For each point $p \in \mathcal{P}_i$, the scale parameter is applied as $p^W = \hat{s}_i R_i p^C + t_i$, to transforming the point from camera coordinates $C$ to the world coordiate $W$. This scale parameterization of each anchor points group allows online depth-scale adjustment during the training time, to mitigate inherent scale-ambiguity problem of monocular depth images.

In complex ground scenes, SfM or LiDAR point cloud data often contains missing 3D structures due to frequent occlusion and lack of observation. This leads to significant drifts in 3DGS algorithms when sufficient multi-view observations are unavailable, caused by degenerate cloning through ADC in 3DGS (Fig. \ref{fig:degen_comparison}). Our monocular initialization produces pixel-aligned, fully covered initial point clouds, while addressing per-view depth-scale ambiguity through the proposed parameterization and scale-consistent depth loss that is further explained in the Sec. \ref{sec:method_C}.

\subsection{Anchored Gaussian Splat Generation}
\label{sec:method_B}
\noindent\textbf{Residual-Form Gaussian Decoder} We define our Residual-Form Gaussian Decoders to generate $k$ Gaussian splats from each anchor. Each anchor point $p_j \in \mathcal{P}$ is associated with a feature descriptor $f_j \in \mathbb{R}^{32}$, nominal Gaussian splat attributes, including position $\bar{\mu}_j \in \mathbb{R}^3$, color $\bar{c}_j \in \mathbb{R}^3$, opacity $\bar{o}_j \in \mathbb{R}$, and scaling $\bar{s}_j \in \mathbb{R}^3$ for covariance composition. For clarity, we slightly modify our notation such that $\bar{\mu}_j$ represents the position of the anchor point, while $p_j$ denotes the point itself. We set nonominal values of covariance related rotation $\bar{r}_j$ as identity quaternions. Our residual decoders $F_{\mu}, F_o, F_c, F_s, F_r$ are defined for each Gaussian attribute $\alpha \in \{\mu, o, c, s, r\}$ lightweight 2-layer Multi-Layer Perceptron (MLP) structures. During training and rendering, the decoders generates $\textit{on-the-fly}$ the residuals $\Delta {\alpha}_i$ from nominal attribute $\bar{\alpha}_i$ stored in each anchor $p_j$, as follows:

\begin{equation}
\{\Delta \alpha_0, \Delta \alpha_1, ... ,\Delta \alpha_{k-1}\}  = F_\alpha (f_j)
\end{equation}

The use of residual-form decoders along with nominal attributes $(\bar{\mu}_j, \bar{c}_j, \bar{o}_j, \bar{s}_j)$ enables faster training of decoders and direct initialization of splat attributes, offering significant advantages over other anchored Gaussian methods \cite{ververas2024sags, lu2024scaffoldgs}.



\vspace{0.5mm}
\noindent\textbf{Anchored Gaussian Generation} $k$ child Gaussian splats are spawned from each anchor $p_j$ by combining the decoded residual attributes $\{\alpha\}_{1:k}$ and nominal attributes $\bar{\alpha}_j$. This is expressed as:

\begin{align}
\mu_{1:k} &= \bar{\mu}_{j} + \Delta \mu_{1:k} \\
c_{1:k} &= \bar{c}_{j} + \Delta c_{1:k} \\
s_{1:k} &= \bar{s}_{j} \cdot \Delta s_{{1:k}} \\
o_{1:k} &= \bar{o}_{j} + \Delta o_{{1:k}} \\ 
\end{align}

As shown in Fig. \ref{fig:color_init}, our residual attribute structure enables direct initialization of splat attributes $\alpha_{1:k}$ (e.g. color) by incorporating the reference value $\bar{\alpha}$. This approach accelerates the training of decoders, as they only need to learn the deviations from the reference value. By addressing one of the main weaknesses of the anchor-decoder scaffold structure \cite{lu2024scaffoldgs, ververas2024sags}, our method improves both the training efficiency and robustness of the original framework.


\begin{figure*}[t]
    \centering
    \includegraphics[width=0.95\linewidth]{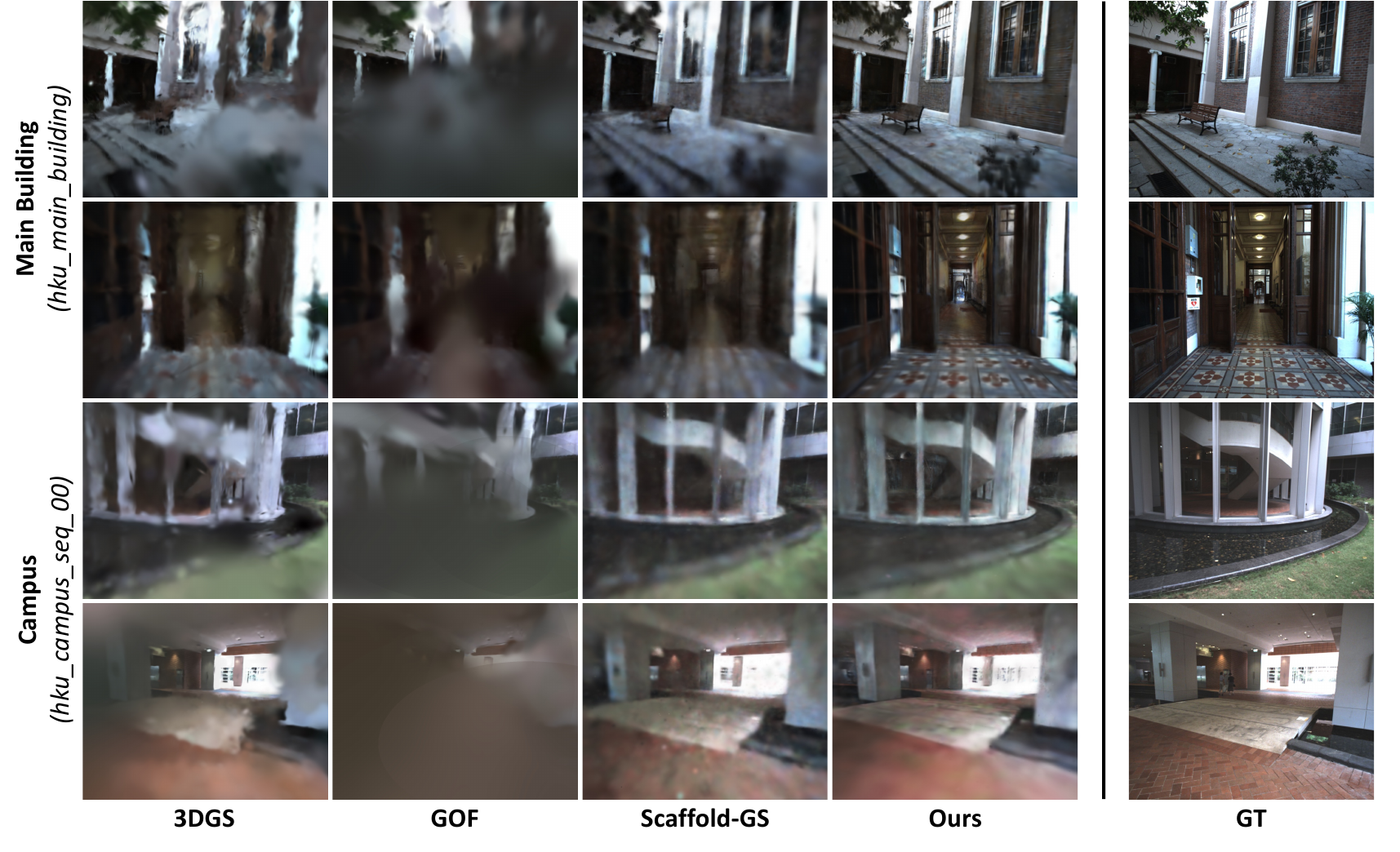}
    \vspace*{-5mm}
    \caption{ 
    Qualitative comparison on two scenes from the R$^{3}$LIVE dataset. Non-anchored methods, such as 3DGS \cite{kerbl20233d} and GOF \cite{gof}, exhibit significant splat drift in the absence of dense multi-view information in sparsely captured scenes. In contrast, both Scaffold-GS \cite{lu2024scaffoldgs} and our method demonstrate robust performance due to their use of anchored splatting. Our approach delivers sharper and more accurate results, attributed to fast training from the direct initialization of splat attributes and dense, pixel-aligned anchor initialization from monocular depth estimation.
    \label{fig:qual_figure}
    }
\end{figure*}

\subsection{Training from Rendering Losses} \label{sec:method_C}
The generated Gaussian splats provide an explicit 3D scene representation that can be rendered into novel view color images and depth images, $\hat{I}, \hat{D}$ using the tile-based rasterizer. We use the color image $I$  and monocular depth image $D$ as supervision during training.

\vspace{0.5mm}
\noindent\textbf{Scale-Consistent Depth Loss} Monocular depth images $D$ inherently contain scale ambiguity, and therefore need to be calibrated with adequate scale parameters when they are used for depth supervision. Unlike previous approaches that employ depth losses based on scale-invariant Pearson Correlation \cite{dnsplatter}, we define our depth loss term with a depth-scale parameter $\hat{\lambda}_i$ embedded for each monocular depth $D_i$, as follows:

\begin{equation}
\mathcal{L}_\text{depth} = \sum_{i =1} ^n \log {|| 1 + (\hat{\lambda}_i  D_i - \hat{D_i}) ||_2}
\end{equation}

Note that this depth-scale parameter $\hat{\lambda}_i$ differs from the scale parameter $\hat{s}_i$ introduced in our anchor depth-scale parameterization. Specifically, $\hat{s}_i$ allows to each initialized per-view anchor group to adjust its scale toward the reference view, while $\hat{\lambda}_i$ corrects the monocular depth scale ambiguity during the loss calibration. 

\noindent\textbf{Full Loss Design} In addition to our proposed scaled-depth loss, the full loss function consists of a photometric loss $\mathcal{L}_\text{photo}$, a volumetric regularization loss $\mathcal{L}_\text{vol}$ \cite{Lombardi21}, and an anisotorpic regularization loss $\mathcal{L}_\text{aniso}$ \cite{xie2023physgaussian}. For completeness, we list these loss functions below:

\begin{align}
\mathcal{L}_\text{photo} &= w \cdot \text{D-SSIM}(I_i, \hat{I}_i) + (1 - w) \cdot || I_i - \hat{I}_i||_1 
\end{align}
In this equation, $\mathcal{L}_\text{photo}$ represents a combination of $\mathcal{L}_1$ loss and D-SSIM loss \cite{kerbl20233d}, where SSIM stands for the Structural Similarity Index Measure . The weight parameter $w$ controls the balance between the two loss components.

\begin{align}
\mathcal{L}_\text{vol} &= \sum_{p \in \mathcal{P}} \text{Prod}(s_i) \\
\mathcal{L}_{aniso} &= \frac{1}{|\mathcal{P}|}\sum_{p \in \mathcal{P}} \text{max}\{\text{max}(s_p)/\text{min}(s_p), r\}-r
\end{align}

Both $\mathcal{L}_\text{vol}$ and $\mathcal{L}_\text{aniso}$ are applied to splats $\mathcal{P}$ to regularize their shape. Here, Prod means the multiplication of covariance-related Gaussian scales \cite{xie2023physgaussian}. The overall loss function is formulated as:

\begin{equation}
\mathcal{L} = \lambda_p \mathcal{L}_\text{photo} + \lambda_s \mathcal{L}_\text{scale} + \lambda_d \mathcal{L}_\text{depth} + \lambda_u \mathcal{L}_\text{aniso}
\end{equation}

Here, $\lambda_p, \lambda_s, \lambda_d$, and $\lambda_u$ represent the weighting factors assigned to each corresponding loss term.


\section{Experiment}

We evaluate our method on four challenging ground-view scenes: one indoor and one outdoor scene each from the R$^{3}$LIVE odometry dataset \cite{lin2022r3live} and the Tanks and Temples dataset \cite{Knapitsch2017tnt}. To demonstrate the performance of our method, we report rendering metrics including PSNR, SSIM \cite{wang2004ssim}, and LPIPS \cite{zhang2018lpips}, which are widely used in neural rendering benchmarks \cite{wang2023f2nerf, kerbl20233d}. Our algorithm utilized no input point clouds in any of the scenes, while we used LiDAR point cloud for R$^{3}$LIVE and SfM point cloud for Tanks and Temples dataset. 

For effective comparison and analysis, we categorized existing 3DGS variants into four types according to the the characteristics of regularization or information that they utilize: (1) \textbf{Geometric} 3DGS methods include 2D Gaussian Splatting (2DGS) \cite{2dgs} and Gaussian Opacity Field (GOF) \cite{gof}. To align the splats with the actual geometry, 2DGS constraint splats to be flat and GOF applies depth distortion loss. (2) \textbf{Sequential} methods, including Colmap-Free 3D Gaussian Splatting (CF-3DGS) \cite{Colmap-free3DGS} and MonoGS \cite{MonoGS}, process each frame sequentially to fully exploit local information and refine image poses. (3) \textbf{Anchored} methods, such as Scaffold-GS \cite{lu2024scaffoldgs} and our approach, generate splats anchored to points with restricted movement in 3D space. Finally, (4) \textbf{Baseline} include the original 3DGS \cite{kerbl20233d} and Mip-Splatting \cite{yu2024mip}.

\subsection{Rendering evaluation on R$^{3}$LIVE dataset}
R$^{3}$LIVE dataset is a publicly available odometry dataset captured by a hand-held device with a 15 Hz camera, 200 Hz Inertial Measurement Unit (IMU), and 10 Hz Livox Avia LiDAR sensor. It includes diverse indoor and outdoor scenes from the campuses of HKU and HKUST. 
Unlike typical neural rendering datasets with object-centric views or structured viewing patterns \cite{Knapitsch2017tnt, Turki2022meganerf, barron2022mipnerf360}, R$^{3}$LIVE dataset captures many complex indoor and outdoor structures with free trajectory patterns.

We process IMU, LiDAR, and Image data using R$^{3}$LIVE \cite{lin2022r3live} multi-sensor odometry pipeline to generate pose-tagged image sequences. To synchronize pose estimation with image time stamps, we slightly modified the R$^{3}$LIVE odometry implementation. However, it still inevitably introduces pixel-level errors due to sensor fusion and inaccurate extrinsic calibration. To avoid redundancy from high-frame rate images, we subsample the images by selecting every 10th frame from the dataset.

As shown in Table \ref{table:r3live_psnr}, our method outperforms all state-of-the-art 3DGS variants in terms of rendering performance. Notably, all algorithms except the Anchored variants exhibit significantly lower performance in the main building scene, which presents considerable challenges due to the complexity of its narrow corridors and hallways. In both scenes, our algorithm achieves state-of-the-art rendering performance even without using initial LiDAR point clouds. Monocular depth based scene initialization delivers much better or comparable performance to LiDAR based intiailzation in both scenes. This is largely due to the inherent incompleteness of LiDAR point clouds in such complex environments. Our results demonstrate that effectively integrating monocular depths can be more beneficial in these challenging scenes, directly generating pixel-aligned and complete point clouds.

\begin{table}[t]
    \begin{center}
    \caption{Results on $R^{3}$LIVE Dataset}
    \label{table:r3live_psnr}
    \resizebox{0.99\linewidth}{!}{
    \begin{tabular}{l|l|l| ccc| ccc}
        \toprule
        \multirow{2}{*}{Method} & \multirow{2}{*}{Type} & \multirow{2}{*}{Input} & \multicolumn{3}{c|}{Main Building} & \multicolumn{3}{c}{Campus} \\
        & & & PSNR $\uparrow$ & SSIM $\uparrow$ & LPIPS $\downarrow$  & PSNR $\uparrow$ & SSIM $\uparrow$ & LPIPS $\downarrow$ \\
        \midrule
         \multicolumn{1}{l|}{3DGS \cite{kerbl20233d}} &\multicolumn{1}{l|}{Baseline}  &\multicolumn{1}{l|}{LiDAR} &15.84 & 0.684 & \multicolumn{1}{c|}{0.497} &   21.60 & 0.760 & 0.364  \\
         
          \multicolumn{1}{l|}{Mip-Splatting \cite{yu2024mip}} &\multicolumn{1}{l|}{Baseline}& \multicolumn{1}{l|}{LiDAR} &13.58 & 0.630 & \multicolumn{1}{c|}{0.499} &   15.89 & 0.679 & 0.449  \\
        \midrule
        \multicolumn{1}{l|}{CF-3DGS \cite{Colmap-free3DGS}}  & \multicolumn{1}{l|}{Sequential}& \multicolumn{1}{l|}{Mono} &-- & -- & \multicolumn{1}{c|}{--} &  --  & -- & -- \\
        \multicolumn{1}{l|}{MonoGS \cite{MonoGS}} &  \multicolumn{1}{l|}{Sequential}&\multicolumn{1}{l|}{Mono} &11.60 & 0.486 & \multicolumn{1}{c|}{0.557} &  16.06 & 0.600 & 0.527   \\
        \midrule
        \multicolumn{1}{l|}{GOF \cite{gof}} &\multicolumn{1}{l|}{Geometric}&\multicolumn{1}{l|}{LiDAR} &  15.53 & 0.676 & \multicolumn{1}{c|}{0.502} & 19.78 & 0.734 & 0.478 \\
        \multicolumn{1}{l|}{2DGS \cite{2dgs}} &\multicolumn{1}{l|}{Geometric}& \multicolumn{1}{l|}{LiDAR} & 15.72 & 0.674 & \multicolumn{1}{c|}{0.507} & 20.67 & 0.743 & 0.418 \\
        \midrule
         \multicolumn{1}{l|}{Scaffold-GS \cite{lu2024scaffoldgs}}   & \multicolumn{1}{l|}{Anchored} &\multicolumn{1}{l|}{LiDAR} & 17.01 & 0.697 & \multicolumn{1}{c|}{0.495} & 20.94 & 0.756 & 0.419 \\
        \multicolumn{1}{l|}{\textbf{Ours w/ mono}}  &\multicolumn{1}{l|}{Anchored}& \multicolumn{1}{l|}{Mono} &\textbf{17.27} & \textbf{0.703} & \multicolumn{1}{c|}{\textbf{0.470}} &  \textbf{22.98} & \textbf{0.774} & \textbf{0.365} \\
        \bottomrule
    \end{tabular}
    }
    \end{center}
\end{table}

\begin{table}[t]
    \begin{center}
    \caption{Results on Tanks and Temples dataset}
    \label{table:tnt_psnr}
    \resizebox{0.99\linewidth}{!}{
    \begin{tabular}{l|l|l| ccc| ccc}
        \toprule
        \multirow{2}{*}{Method} & \multirow{2}{*}{Type} & \multirow{2}{*}{Input} & \multicolumn{3}{c|}{Main Building} & \multicolumn{3}{c}{Campus} \\
        & & & PSNR $\uparrow$ & SSIM $\uparrow$ & LPIPS $\downarrow$  & PSNR $\uparrow$ & SSIM $\uparrow$ & LPIPS $\downarrow$ \\
        \midrule
        \multicolumn{1}{l|}{3DGS \cite{kerbl20233d}} &\multicolumn{1}{l|}{Baseline}& \multicolumn{1}{l|}{SfM} & 15.92 & 0.689 & \multicolumn{1}{c|}{0.396} & 16.86 & 0.639 & 0.435 \\
        \multicolumn{1}{l|}{Mip-Splatting \cite{yu2024mip}}&\multicolumn{1}{l|}{Baseline} &\multicolumn{1}{l|}{SfM} &15.44 & 0.683 & \multicolumn{1}{c|}{0.366} & 16.39 & 0.646 & 0.417 \\
        \midrule
        \multicolumn{1}{l|}{CF-3DGS \cite{Colmap-free3DGS}} &\multicolumn{1}{l|}{Sequential}& \multicolumn{1}{l|}{Mono} &--  & -- & \multicolumn{1}{c|}{--} & 15.53  & 0.614 & 0.510 \\
        \multicolumn{1}{l|}{MonoGS \cite{MonoGS}}&\multicolumn{1}{l|}{Sequential} &\multicolumn{1}{l|}{Mono} & 9.78 & 0.453 & \multicolumn{1}{c|}{0.637} & 12.01  & 0.488 & 0.591 \\
        \midrule
        \multicolumn{1}{l|}{GOF \cite{gof}} &\multicolumn{1}{l|}{Geometry}& \multicolumn{1}{l|}{SfM} &15.49 & 0.680 & \multicolumn{1}{c|}{0.377} & 16.56 & 0.636 & 0.459 \\
        \multicolumn{1}{l|}{2DGS \cite{2dgs}} &\multicolumn{1}{l|}{Geometry}& \multicolumn{1}{l|}{SfM} &16.57 & 0.705 & \multicolumn{1}{c|}{0.382} & 17.20 & 0.636 & 0.455 \\
        \midrule
        \multicolumn{1}{l|}{Scaffold-GS \cite{lu2024scaffoldgs}} &\multicolumn{1}{l|}{Anchored}&\multicolumn{1}{l|}{SfM} & \textbf{17.12} & \textbf{0.719} & \multicolumn{1}{c|}{\textbf{0.345}} & \textbf{17.42} & \textbf{0.677} & \textbf{0.413} \\
        \multicolumn{1}{l|}{\textbf{Ours w/ mono}} &\multicolumn{1}{l|}{Anchored}& \multicolumn{1}{l|}{Mono} &15.70 & 0.682 & \multicolumn{1}{c|}{0.442} & 16.66 & 0.641 & 0.462 \\
        \bottomrule
    \end{tabular}
    }
    \end{center}
\end{table}

\subsection{Rendering evaluation of Tanks and Temples dataset}
We also validate our method on the Tanks and Temples dataset, a widely recognized benchmark for neural rendering evaluation. We select the courthouse scene and the meeting room scenes as they are geometrically most challenging in the benchmarks \cite{gof}. Similar to the evaluation of R$^{3}$LIVE dataset, we subsample every 10th images. Instead of LiDAR point clouds, we directly use SfM PCD for the other variants.

Unlike the R$^{3}$LIVE odometry dataset, the trajectories in this dataset follow object-centric or circular patterns, providing relatively dense observations to each part of scene. Due to this reason, our method does not achieve the best performance. In these viewing patterns, the initial point cloud has less impact on 3DGS algorithms, as splat cloning and splitting are effectively guided by salient multi-view photometric gradients. It has even been shown that random initialization can yield plausible results in such cases \cite{jung2024relaxing}. As monocular depth usually contains inevitable inner distortion that can not be adjusted by scale factor, anchoring on these depths can be detrimental when enough photometric information is given. Nonetheless, the best-performing algorithm in this scenario is the anchor-based Scaffold-GS \cite{lu2024scaffoldgs}, validating our analysis of the degeneracy patterns associated with different algorithm types (Fig. \ref{fig:degen_comparison}).

As shown in Table. \ref{table:tnt_psnr}, our method still shows comparable performance to other 3DGS methods. In this sense, Our method achieves a suitable balance between robustness to limited multi-view information and high performance in densely captured datasets, demonstrating the most stable performance across all the datasets from R$^{3}$LIVE and Tanks and Temples datasets.

\subsection{Ablation Studies}
\begin{figure}[t]
    \subfigure[]{\includegraphics[width=0.63\linewidth]{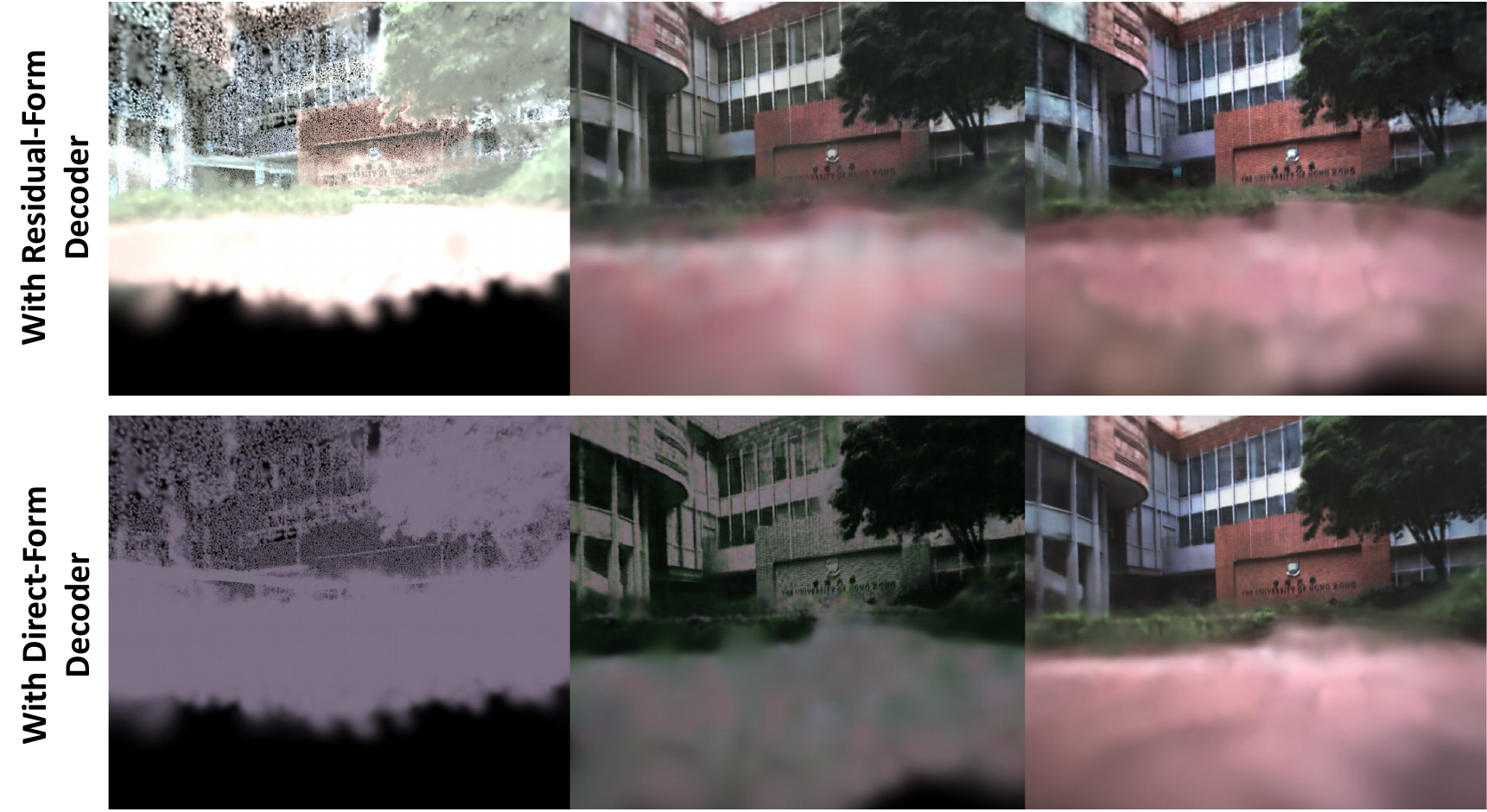}\label{fig:color_init}}
    \subfigure[]{\includegraphics[width=0.33\linewidth]{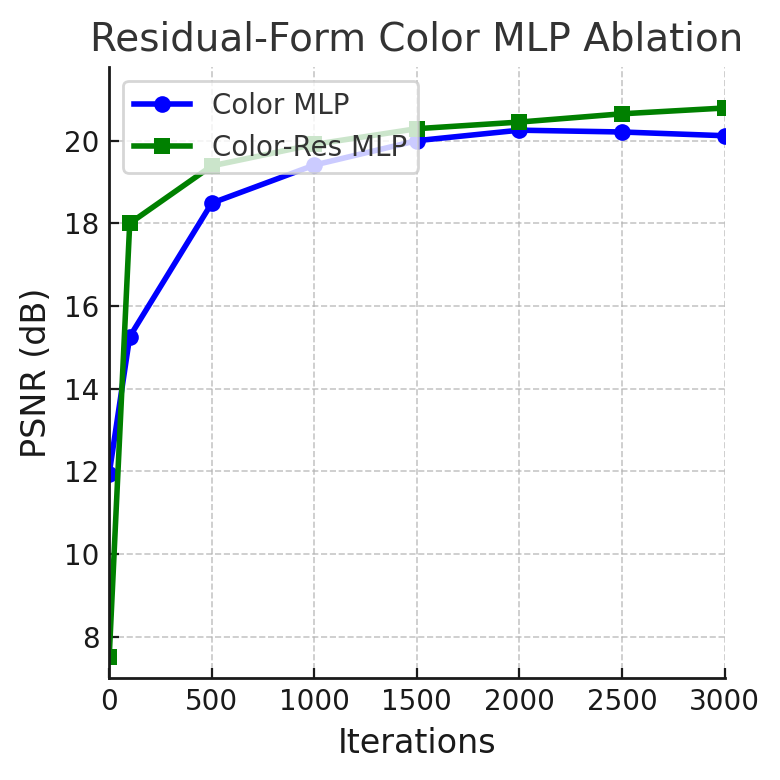}\label{fig:color_init_plot}}  
    \vspace{-3mm}
    \caption{(a)  \textbf{With Residual-Form Gaussian Decoder (top)} , only residual from nominal color is estimated and trained by the decoder, allowing direct color initialization and fast training. \textbf{Direct-Form Gaussian Decoder (bottom)} \cite{lu2024scaffoldgs, ververas2024sags} does not allow color initialization due to its \textit{on-the-fly} decoding scheme. (b) Rendering Performance (PSNR) ablation between Direct-Form Color MLP and Residual-Form Color MLP.}
\end{figure}

We evaluated our depth calibration framework and residual-form Gaussian MLP in Table \ref{ablatation_table}. As shown in the ablation results, each proposed module contributes to the overall improvement in rendering performance. Additionally, our residual-form Gaussian decoder enables fast initialization of Gaussian attributes, as illustrated in Fig. \ref{fig:color_init}. For this ablation, we used a direct-form MLP that generates color attributes from features, similar to \cite{lu2024scaffoldgs, ververas2024sags}. Compared to this direct-form MLP decoder, our proposed decoder significantly accelerates the training process (Fig. \ref{fig:color_init_plot}).

\begin{table}[t]
    \begin{center}
    \caption{Ablation of each proposed module}
    \label{ablatation_table}
    \resizebox{0.7\linewidth}{!}{
    \begin{tabular}{ccc ccc}
        \toprule
        depth cal. & res. MLP &  PSNR $\uparrow$ & SSIM $\uparrow$ & LPIPS $\downarrow$ \\
        \midrule
        &\multicolumn{1}{l|}{}  & 16.84 & 0.687 & 0.475 \\
        \midrule
        \checkmark & \multicolumn{1}{l|}{} & 16.91 & 0.693 & 0.467 \\
        \midrule
        &  \multicolumn{1}{l|}{\checkmark } & 17.03 & 0.688 & 0.465 \\
        \midrule
        \checkmark & \multicolumn{1}{l|}{\checkmark } & 17.27 & 0.703 & 0.470 \\
        \bottomrule
    \end{tabular}
    }
    \end{center}
\end{table}

\section{Conclusions, Limitations, and Future Work}

In this paper, we presented Mode-GS, a novel 3DGS algorithm designed for robust neural rendering from ground-robot trajectory datasets. Our algorithm introduces a practical rendering pipeline for ground-view robot datasets, utilizing easily obtainable odometry poses and operating in a point-cloud-free setting. However, our approach is less effective in scenarios where extensive multi-view data is available, such as in densely captured, object-centric datasets. Future work will focus on developing a hybrid approach that integrates our method with non-anchored splats to achieve optimal performance.
\bibliographystyle{IEEEtran}
\bibliography{references}

\addtolength{\textheight}{-12cm}   






\end{document}